\documentclass[journal,onecolumn]{IEEEtran}
\ifCLASSINFOpdf
\else
\fi
\usepackage{graphicx}
\usepackage{amsmath}
\usepackage{amssymb}
\usepackage{amsthm}
\usepackage{booktabs}
\usepackage{algorithm}
\usepackage{algorithmic}
\usepackage{subfigure}

\begin{document}
%
\title{Disentanglement Then Reconstruction: Learning Compact Features for Unsupervised Domain Adaptation}
%
%
%


\author{Lihua~Zhou, Mao~Ye$^*$, ~\IEEEmembership{Member,~IEEE,}
Xinpeng~Li,
~Ce~Zhu, ~\IEEEmembership{Fellow,~IEEE,}
~Yiguang~Liu,~\IEEEmembership{Member,~IEEE}
and ~Xue~Li, ~\IEEEmembership{Member,~IEEE}

\thanks{Lihua~Zhou, Mao Ye and Xinpeng Li are with the School of Computer Science and Engineering, University of Electronic Science and Technology of China, Chengdu 611731, P.R. China (e-mail: cvlab.uestc@gmail.com).}
\thanks{Ce Zhu is with the School of Information and Communication Engineering, University of Electronic Science and Technology of China, Chengdu 611731, P.R. China (e-mail: eczhu@uestc.edu.cn)}
\thanks{Yiguang Liu is with the Vision and Image Processing Laboratory, School of Computer Science, Sichuan University, Chengdu 610065, P.R. China.(email: lygpapers@aliyun.com)}
\thanks{Xue Li is with the School of Information Technology and Electronic Engineering, The University of Queensland, Brisbane, QLD 4072, Australia.(email: xueli@itee.uq.edu.au)}
\thanks{*corresponding author}}

\maketitle

\begin{abstract}
  Recent works in domain adaptation always learn domain invariant features
  to mitigate the gap between the source and target domains by adversarial methods. The category information are not sufficiently used which causes the learned domain invariant features are not enough discriminative. We propose a new domain adaptation
  method based on prototype construction which likes capturing data cluster centers.
  Specifically, it consists of two parts: disentanglement and reconstruction.
  First, the domain specific features and domain invariant features are
  disentangled from the original features. At the same time,
  the domain prototypes and class prototypes of both domains are estimated.
  Then, a reconstructor is trained by reconstructing the original features
  from the disentangled domain invariant features and domain specific features.
  By this reconstructor, we can construct prototypes for the original features
  using class prototypes and domain prototypes correspondingly. In the end,
  the feature extraction network is forced to extract features close to
  these prototypes. Our contribution lies in the technical use of
  the reconstructor to obtain the original feature prototypes which helps to learn compact and discriminant features.
  As far as we know, this idea is proposed for the first time.
  Experiment results on several public datasets confirm the state-of-the-art
  performance of our method.
\end{abstract}


%
\IEEEpeerreviewmaketitle

\section{Introduction}
%
%
%
%
\IEEEPARstart{R}{ecently}, with the development of deep learning,
machine learning has made significant breakthroughs
 in various fields, such as image classification \cite{russakovsky2015imagenet},
 but this breakthrough is based on a large number of labeled data. Unfortunately,
 in the practical applications, we usually only have unlabeled data or very little
 labeled data. Therefore, it's a strong motivation to build an effective model for the target domain by using the available labeled data of source domains.
 However, due to domain shift \cite{torralba2011unbiased}, the model trained in source domain will lead to performance degradation in target domain.
 In order to solve this problem, domain adaptation has attracted a lot attentions recently, which usually seek to minimize both the source domain task error and distribution discrepancy between source and target domain.
 In this work, we focus on the
 unsupervised domain adaptation which aims to transfer
 knowledge from a label rich source domain to an unlabeled target domain \cite{pan2010a}.

 \begin{figure*}[t]
  \begin{center}
     \includegraphics[width=0.65\linewidth]{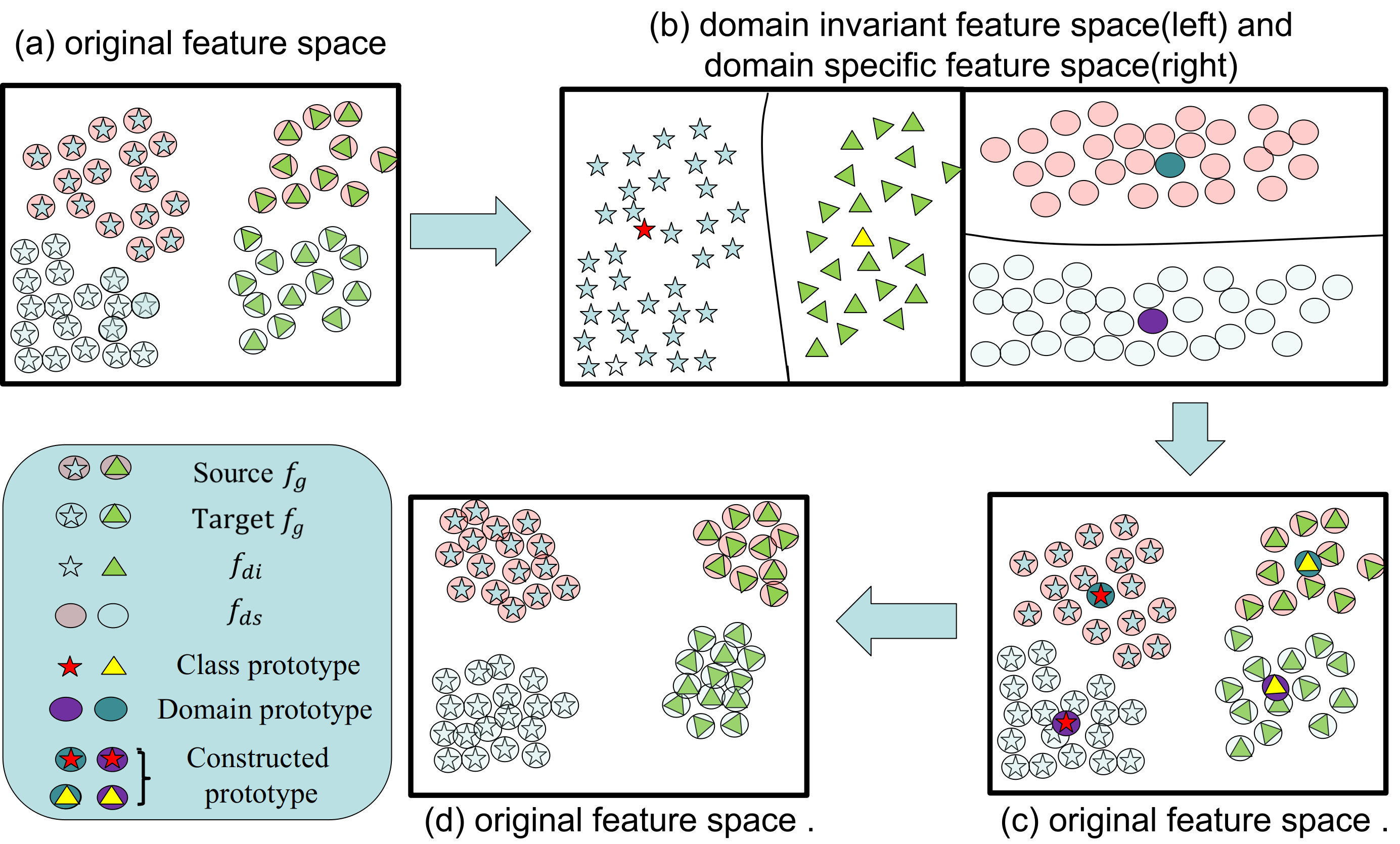}
  \end{center}
     \caption{The process of learning compact features. As shown in the legend, pentagrams and triangles represent the domain invariant features with different categories and the circles with different color represent the domain specific features from different domains respectively. (a) The original features are scattered.
     (b) The disentangled domain invariant features $f_{di}$ and domain specific features $f_{ds}$. The corresponding class and domain prototypes are shown.
     (c) The original feature prototypes are constructed by using class and domain prototypes.
     (d) More compact features can be learned by forcing the features close to the prototypes. Thus the domain invariant features will be more compact and discriminative.
     }
  \end{figure*}

The technical routes for domain adaptation can be roughly divided into two strategies. The first strategy is based on statistic moment matching, which models the discrepancy between the source and target domains as loss function, such as Maximum Mean Discrepancy (MMD). Minimizing the discrepancy mitigates the gap between the source and target domains. The methods represented by such strategy are TCA\cite{pan2011domain}, JDA\cite{Long2013TransferFL} in traditional machine learning; and DAN\cite{long2015learning}, JAN\cite{long2017deep} and AFN\cite{xu2019larger} using deep learning approach and so on.
Another strategy is using adversarial learning technique, which forces feature extraction network to extract domain invariant feature by confusing domain discriminator in an adversarial learning paradigm. The methods represented by such strategy are DANN\cite{ganin2016domain-adversarial}, CDAN\cite{long2018conditional} and MCD\cite{saito2018maximum}, etc.
At present, adversarial domain adaptation is the mainstream research method for domain adaptation. Although adversarial domain adaptation can enhance the feature
alignments between the source and target domains, there exists some loss of feature discriminability \cite{chen2019transferability}.

It is worth noting that most of previous unsupervised domain adaptation methods considered only global alignment but ignored category information.
Specifically, for statistic moment matching, we usually
calculate the divergence of samples in the whole domain
without considering the category information;
for the adversarial learning based approach,
domain discriminator also only concerns the domain category.
This makes the learned features are not enough discriminative, which leads to the performance degradation\cite{Motiians,Hoffman,luo}.
Recently, some methods use class prototypes to save category information which has been used in supervised domain adaptation\cite{Motiians,luo} and unsupervised domain adaptation \cite{xie2018learning}.
All methods introduce two sets of class prototypes to represent the category information from the source and target domains respectively.
In our method, we also introduce the class prototype to save the category information,
but we only introduce a set of class prototypes to represent the category information of all domains which can better mitigate the gap between two domains.

Motivated by the work in \cite{peng2019domain},  we put forward a hypothesis that the feature can be considered as a non-linear combination
of domain specific feature $f_{ds}$ and domain invariant feature $f_{di}$ as the following,
\begin{equation}
   G(img) = f_{g} =f_{ds} \circ f_{di},
   \label{eq:Sample}
\end{equation}
where $G$ is the feature extraction network, and $\circ$ represents a non-linear combination. Specifically, domain specific feature $f_{ds}$ is the domain related information; while domain invariant feature $f_{di}$ is the intrinsic information related to the category. Traditional approaches try to extract the domain invariant features, and then retrain the source classifier based on these features.
In our method, we further try to disentangle domain specific features to estimate domain prototypes.
The domain prototypes can be used to construct the original feature prototypes, so compared with the method in \cite{peng2019domain}, our method makes full use of domain specific features.

Based on the above analysis, we propose a novel method, named Disentanglement Then Reconstruction (DTR). Disentanglement means domain invariant feature and domain specific feature are extracted from the original feature for both the source and target domains.  Adversarial learning technique is employed to learn a disentangler for extract domain invariant feature by which a linear category classifier is trained at the same time. Another disentangler and a linear domain classifier for domain specific feature can be trained based on cross-entropy loss. We can get the corresponding class and domain prototypes according to these linear classifiers respectively.  Reconstruction consists of two parts. First, a reconstructor  is learned using the disentangled domain invariant features and domain specific features.
Then, the original feature prototypes can be obtained by this reconstructor by using class prototypes and domain prototypes. Finally, the feature extraction network is retrained by forcing the features close to the corresponding prototypes.
The main contributions of our method can be summarized as follows:

1) We propose a new technical route for unsupervised domain adaptation. First, domain prototypes and class prototypes are estimated by the domain features and class features in the disentanglement process, then the original feature prototypes are constructed in the reconstruction process, and finally the original feature prototypes is used to supervise feature extraction network to learn more compact features.

2) The constructed prototypes are used to train the feature extraction network. Different from the traditional approaches, it is first proposed that the technical use of the reconstructor to obtain the prototypes to update classifiers. Thus category information are sufficiently used.

3) The experiments on three public datasets are conducted. Our method works very well.
Thorough parameter experiments and feature visualizations also illustrate the advantages and robustness of our method.

\section{Related work}

Domain adaptation has attracted a lot of attentions recently.
The methods in domain adaptation can be divided into
two  categories: statistic moment matching and
adversarial domain adaptation. Statistic moment matching
reduces the discrepancy of feature distribution between
the source and the target domains by minimizing a
clearly defined statistical distance, such as Maximum Mean Discrepancy (MMD).
Early domain adaptation methods are researched mainly with this kind of strategy in the shallow regime\cite{pan2011domain,Long2013TransferFL,saenko2010adapting,gong2012geodesic,duan2012domain,zhang2013domain,wang2014flexible}.
With the development of deep learning,  \cite{yosinski2014how}
proves that deep networks can learn more transferable feature, which has also been proved in LRCT\cite{ren2020generalized} that CNN-based features can improve performance better than traditional features through experiments. After that,
researchers start to study domain adaptation with deep networks as the basic framework
\cite{long2015learning,long2017deep,tzeng2014deep,Motiian,Pinheiro,wu2019joint,MaC,ren2019heterogeneous}.
DDC in \cite{tzeng2014deep} proposes to
minimize Maximum Mean Discrepancy (MMD) for the bottleneck features of the last layer.
DAN in \cite{long2015learning}  proposes
to minimize Multi-kernel MMD for the bottleneck features of the last three layers.
JAN in \cite{long2017deep} further extends JDA in \cite{Long2013TransferFL} with deep learning as a framework
by adapting the joint distribution of the source and target domains.

\begin{figure*}[t]
  \begin{center}
     \includegraphics[width=0.8\linewidth]{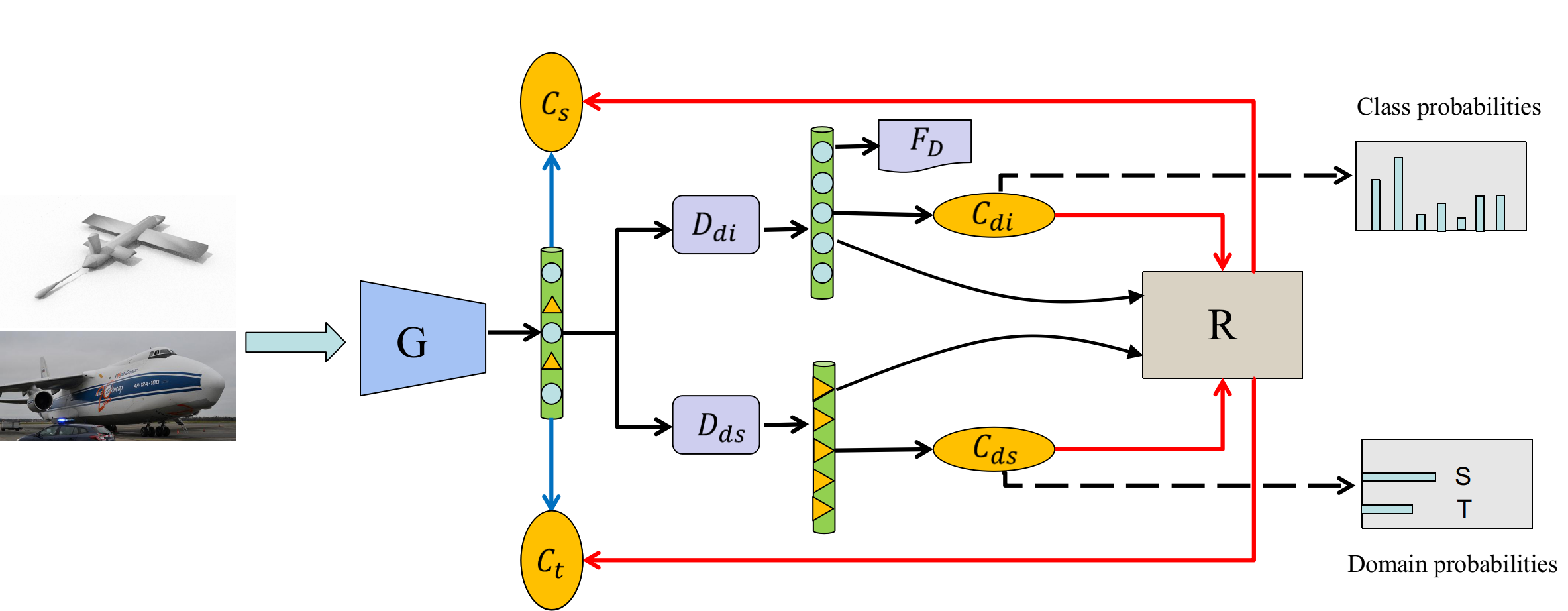}
  \end{center}
     \caption{An overview of our method.
     For any image, we first use the feature extraction network $G$ to extract
     the original features, then use the two disentanglers $D_{di}$ and $D_{ds}$
     to disentangle the domain invariant features and domain specific features respectively,
   at the same time, we use two classifiers $C_{di}$ and $C_{ds}$ to evaluate the class prototypes and domain prototypes respectively(black line).
   Then we train the reconstruction network to reconstruct the original features by using the disentangled features.
   Then we use the reconstruction network to construct the original prototypes by using evaluated prototypes, and send them into  classifiers $C_{s}$ and $C_{t}$ correspondingly(red line).
   Finally, we use classifiers $C_{s}$ and $C_{t}$ to supervise the feature extraction network $G$ to learn more compact features(blue line).
     }
\end{figure*}

Inspired by GAN\cite{NIPS2014_5423}, adversarial domain adaptation introduces a domain discriminator
to discriminate features from source and target domains and
force the feature extraction network to confuse the domain discriminator
in an adversarial learning paradigm to learn domain invariant feature\cite{ganin2016domain-adversarial,long2018conditional,saito2018maximum,tzeng2015simultaneous,tzeng2017adversarial}.
DANN in \cite{ganin2016domain-adversarial} is a pioneering work on adversarial domain adaptation methods,
it combines domain adaptation with GAN for the first time. On this basis,
ADDA in \cite{tzeng2017adversarial} trains two feature extraction
networks to extract feature to confuse the domain discriminator.
MCD in \cite{saito2018maximum} uses two classifiers as domain discriminator and plays a min-max game
between one feature extraction network and two classifiers.
Different from DANN, CDAN in \cite{long2018conditional} trains domain discriminator on the multilinear map of category classifier prediction and domain invariant feature.

Generally speaking, traditional methods in domain adaptation is easy to make the learned features not discriminative because it does not consider the category information.
The main reason why category information is not considered is the lack of label information of the target domain.
Therefore, in order to solve this problem, many methods use pseudo label to replace the label information of the target domain\cite{Long2013TransferFL,xie2018learning,bruzzone2010domain,Minmin,saito2017asymmetric,sener2016learning,chen2019progressive,zhang2018collaborative}.
JDA in \cite{Long2013TransferFL} uses pseudo label to match conditional distribution by a revised MMD.
ATDA in \cite{saito2017asymmetric} utilizes an asymmetric tri-training strategy to generate pseudo label for target domain to learn discriminative feature.
MSTN in \cite{xie2018learning} assigns pseudo labels to all target samples and uses pseudo labels for semantic alignment besides global alignment.
CAN in \cite{zhang2018collaborative} iteratively selects pseudo-labeled target samples and uses them to retrain model.
PFAN in \cite{chen2019progressive}  aligns the discriminative features across domains progressively by developing an Easy-to-Hard transfer strategy and an adaptive prototype alignment step.
However, these methods  are highly dependent on the correctness of pseudo labels.
Therefore, in order to reduce the dependence on pseudo labels, we propose two improvements. First, we select samples with high confidence pseudo label to train model. Second, we use prototypes and use them to save the category and domain information.

Traditional methods in domain adaptation is easy to get trapped in trivial solutions. Specifically, a global optimal solution is $f(img)=\textbf{0}$ and the optimal objective of global alignment can always be achieved.
And recently many methods introduce autoencoder to avoid trivial solutions\cite{xie2018learning,peng2019domain,cai2019learning,yang2017towards,chen2019domain}; the key to prevent trivial solution lies in reconstruction part.
Like the previous method, we also introduced an autoencoder.
The difference is that our reconstruction loss only optimizes the reconstruction network. We hope that through training reconstruction network, we can fit the nonlinear combination we mentioned earlier.

The method proposed in \cite{peng2019domain} is related to our work. They employ  class disentangler and domain disentangler to remove class irrelevant and domain specific features, and then minimize the mutual information between the disentangled features. This method learns domain invariant features implicitly.
While our method  fully uses  domain invariant  and domain specific features, and learns more compact features which makes our category classifier perform better to a certain extent.
The method proposed in \cite{cai2019learning} is also related to our work. They uses a dual adversarial network to disentangle domain invariant and domain specific features,
and domain invariant feature is required to be independent of domain specific feature, so they can obtain those domain invariant feature without the contamination of the domain information.
Different from them, we use adversarial method to disentangle domain invariant feature and use standard supervised method to disentangle domain specific feature.
Compared with the method in \cite{cai2019learning}, our method is more flexible because we don't have restrictions on domain invariant feature and domain specific feature.

\section{The Proposed Method}
For the problem of unsupervised domain
adaptation, a source domain $\mathcal{D}_{s}=\left\{\left(x_{i}^{s}, y_{i}^{s}\right)\right\}_{i=1}^{n_{s}}$
consists of $n_{s}$ labeled samples, and a target domain $\mathcal{D}_{t}=\left\{\left(x_{i}^{t} \right)\right\}_{i=1}^{n_{t}}$
consists of $n_{t}$ unlabeled samples. Both domains share the same label space $\{1,2,\cdots,K\}$. Due to
the domain shift, the source distribution is different from
the target distribution. Our goal is to learn more compact features using prototypes such that the classification model works well in both domains.

The framework of our method is shown in Fig. 2. The original feature $f_{g}$ is extracted by the feature extraction network $G$.  Then the disentangler $D_{di}$ is adversarially trained with the domain discriminator $F_{D}$ to disentangle domain invariant feature $f_{di}$ from the original feature $f_{g}$. The classifier $C_{di}$ in the source domain is trained based on the domain invariant feature. Another disentangler $D_{ds}$ extracts domain specific feature $f_{ds}$ from $f_{g}$, and the classifier $C_{ds}$ is trained to distinguish the feature $f_{ds}$ from different domains.
It is worth noting that both $C_{di}$ and $C_{ds}$ are linear networks with only weights and no bias,
so the weights of $C_{di}$ and $C_{ds}$ can be seen as class prototypes and domain prototypes respectively\cite{saito2019semi-supervised}.
As with the previous method, the reason for disentangling domain invariant features is to mitigate the gap between the two domains. Thus the knowledge can be transferred from the source domain to target domain.
On this basis, we further propose to disentangle the domain specific features and the reason is that the domain prototypes is estimated by classifying domain specific features so as to save domain information, which can help algorithm construct original feature prototypes to learn more compact original feature.

Next, the reconstructor $R$ is trained by reconstructing the original features
from the domain invariant features and domain specific features.
By this reconstructor, we can construct prototypes for the original features
using class prototypes and domain prototypes correspondingly.
And we set those original feature prototypes to the weights of the linear classifiers $C_{s}$ and $C_{t}$.
In the end,
the feature extraction network $G$ is forced to extract more compact features
by using $C_{s}$ and $C_{t}$.
Generally speaking, prototype is a special feature and a representative point of each class\cite{saito2019semi-supervised}, which can be regarded as "cluster center".
It is worth noting that our reconstruction loss only optimizes the reconstruction network $R$ without back propagation of other networks. This is because we can fit the nonlinear combination mentioned above through such optimization, that is, $R(f_{ds},f_{di}) = f_{ds} \circ f_{di}$.

\subsection{Disentanglement}

First of all, we send images which are randomly selected from source domain and target domain into feature extraction network to obtain original features, $f_{g}=G(img)$.
In the following, our superscripts $s$ and $t$ represent the features from the source domain and the target domain respectively.
According to the hypothesis mentioned above, the original feature is a nonlinear combination of domain invariant feature and domain specific feature. So two disentanglers are used to disentangle domain invariant features and domain specific features from original features respectively. The specific process is as follows.

The disentangler $D_{di}$ is used to disentangle domain invariant features from original features, $f_{di}=D_{di}(f_{g})$.
$D_{di}$ is trained by minimizing the discrepancy between the source and target domains and adversarial method is employed.
A domain discriminator $F_D$ is introduced to distinguish the source domain invariant feature $f_{di}^{s}$ from the target domain invariant feature $f_{di}^{t}$; and the disentangler $D_{di}$ is learned to confuse $F_D$ to minimize the discrepancy between $f_{di}^{s}$ and
$f_{di}^{t}$. The traditional optimization objective\cite{ganin2016domain-adversarial} can be defined as follows:
\begin{equation}
   \begin{aligned}
      \mathcal{E}_{dist}^{\prime} &= \mathbb{E}_{\mathbf{x}_{i}^{s} \sim source} \log \left[F_{D}\left(f_{di}^{s}\right)\right] \\
   &+\mathbb{E}_{\mathbf{x}_{i}^{t} \sim target} \log \left[1-F_{D}\left(f_{di}^{t}\right)\right].
   \end{aligned}
\end{equation}

Recently, CDAN-E in \cite{long2018conditional} proposes a better method to mitigate the gap between two domains compared with traditional method\cite{ganin2016domain-adversarial}.
It finds that the classifier prediction is useful for adversarial learning, because it convey rich discriminative information, so it can better minimize the discrepancy between source and target domain,
which is also mentioned in \cite{mirza2014conditional}.
Therefore, like many recent methods, we also choose to use CDAN-E\cite{long2018conditional} instead of DANN\cite{ganin2016domain-adversarial} as our adversarial training method,
so $\hat{h}$ is firstly calculated by multilinear transformation $\otimes$ of domain
invariant feature $f_{di}$ and the classifier $C_{di}$ predictions $p = C_{di}(f_{di})$:
\begin{equation}
   \hat{h}=f_{di} \otimes p,
   \end{equation}
where multilinear transformation $\otimes$ is defined as outer product of multiple random vectors.
Compared with $f_{di}$, $\hat{h}$ is more discriminative due to the classifier prediction $p$,
 so we employ $\hat{h}$ instead of $f_{di}$. Then the discrepancy between the two domains becomes:
   \begin{equation}
      \begin{aligned}
         \mathcal{E}_{dist} &= \mathbb{E}_{\mathbf{x}_{i}^{s} \sim source} w(x_{i}^{s})\log \left[F_{D}\left(\hat{h}^{s}\right)\right] \\
      &+\mathbb{E}_{\mathbf{x}_{i}^{t} \sim target} w(x_{i}^{t})\log \left[1-F_{D}\left(\hat{h}^{t}\right)\right],
      \end{aligned}
   \end{equation}
where $w(x)=1+e^{-H(p)}$ and $H(p)$ is the entropy of classifier prediction $p$. The purpose of $w(x)$ is to give different weights to different samples according to the classifier predictions $p$, which makes easy-to-transfer samples have greater weight, and then a safer transfer can be achieved.

Furthermore, we also need to minimize the classification errors based on the domain invariant features in the source domain, which is calculated as the following,
\begin{equation}
    \mathcal{E}_{cls}^{s}=\frac{1}{n_{s}} \sum_{i=1}^{n_{s}} \mathcal{L}^{s}\left(C_{di}\left(f_{di}^{s}\right), y_{i}^{s}\right),
   \end{equation}
where $\mathcal{L}^{s}$ is the cross-entropy loss.

For disentangling domain specific feature $f_{ds}$, we introduce a classifier $C_{ds}$, for which, the label of domain specific feature from the source or target domains is set as 1 or 0, respectively. The disentangler $D_{ds}$ and classifier $C_{ds}$ can be trained by minimizing the cross-entropy loss between the domain predictions
and domain labels,
\begin{equation}
    \mathcal{E}_{cls}^{d}=\frac{1}{n_{s}+n_{t}} \sum_{i=1}^{n_{s}+n_{t}} \mathcal{L}^{s}\left(C_{ds}\left(f_{ds}\right), domain\right).
\end{equation}

In summarization, by combining the loss functions (4-6), we have the following training objectives for disentanglement process:

\begin{equation}
   \min _{C_{di},D_{di},C_{ds}, D_{ds},G} \mathcal{E}_{cls}^{s}+\alpha\mathcal{E}_{dist}+\beta\mathcal{E}_{cls}^{d},
\end{equation}
\begin{equation}
   \max _{F_{D}}  \mathcal{E}_{dist},
\end{equation}
where $\alpha$ and $\beta$ are hyperparameters for trading off.
In all experiments $\alpha$ is set to 1 referring to the method proposed in \cite{chen2019transferability}.
$\beta$ is tuned by cross-validation method and set to 0.15 in most experiments.

At the same time, the feature extraction network $G$ is also adjusted accordingly to obtain better domain invariant and domain specific features in this step. According to the linear classifiers $C_{ds}$ and $C_{di}$, the corresponding domain prototypes $(w_d^s,w_d^t)$ and the class prototypes $(w_c^1,\cdots,w_c^K)$, where $K$ represents the number of categories, can be obtained by their weights respectively.

\subsection{Reconstruction}

We propose a two-step reconstruction process, which firstly train a reconstructor to obtain the original feature using domain invariant feature
and domain specific feature, and then construct the original feature prototypes using class prototypes and domain  prototypes.

First, we train a reconstructor $R$ to obtain original feature $f_{g}$ by using the features ($f_{di}$, $f_{ds}$). Assume  $\hat{f_{g}} = R(f_{di}, f_{ds})$, the reconstruction loss is as the following,
\begin{equation}
    \min _{R}\mathcal{E}_{rec}=\left\| \hat{f_{g}}-f_{g}\right\|_{2}^{2}.
\end{equation}
Ideally, through the above optimization of $R$,
we can get the calculation method of the non-linear combination we proposed in Eq. (1), i.e.,
$R(f_{di}, f_{ds}) = f_{ds} \circ f_{di}$.

By this reconstructor $R$, we can construct the prototypes, which can be considered as "cluster center" of original features, for each category in the source and target domains respectively as follows,
\begin{eqnarray}
   &&w_{s}^{i}=R(w_{c}^{i},w_{d}^{s}),\\
   &&w_{t}^{i}=R(w_{c}^{i},w_{d}^{t})
\end{eqnarray}
for $i\in\{1,\cdots,K\}$.  By these prototypes, we can construct the source and target  linear classifiers $C_{s}$ and $C_{t}$, whose weights are $\left[w_{s}^{1}, \cdots, w_{s}^{K}\right]$
and $\left[w_{t}^{1}, \cdots, w_{t}^{K}\right]$ respectively.
The weights of $C_{s}$ and $C_{t}$ are updated by Eqs.(10-11) for each reconstruction interval $r$ iteration. And the weights of $C_{s}$ and $C_{t}$ are not affected by back propagation of any loss function. Usually, a batch of samples is selected from the source and target domains respectively in an iteration. The sensitivity of $r$ will be discussed in Section 4.3.

\begin{algorithm}[t]
   \caption{DTR}
   \label{alg:algorithm}
   \textbf{Input}:$\mathcal{D}_{s}=\left\{\left(\mathbf{x}_{i}^{s}, \mathbf{y}_{i}^{s}\right)\right\}_{i=1}^{n_{s}}$, $\mathcal{D}_{t}=\left\{\left(\mathbf{x}_{i}^{t} \right)\right\}_{i=1}^{n_{t}}$,  iteration batches $N$, reconstruction interval $r$, hyperparameters $\alpha,\beta,\gamma,\theta$\\
   \textbf{Output}:feature extraction network $G$, disentangler $D_{di}$, classifier $C_{di}$
   \begin{algorithmic}[1] 
   \FOR{$n$ = 1:$N$}
   \STATE ($x^{s}$,$y^{s}$) $\gets$ RANDOMSAMPLE($\mathcal{D}_{s}$)
   \STATE ($x^{t}$) $\gets$ RANDOMSAMPLE($\mathcal{D}_{t}$)
   \STATE $f_{g}^{s}$, $f_{di}^{s}$, $f_{ds}^{s}$ $\gets$ $G(x^{s})$, $D_{di}(f_{g}^{s})$, $D_{ds}(f_{g}^{s})$
   \STATE $f_{g}^{t}$, $f_{di}^{t}$, $f_{ds}^{t}$ $\gets$ $G(x^{t})$, $D_{di}(f_{g}^{t})$, $D_{ds}(f_{g}^{t})$
   \STATE Calculate $\mathcal{E}_{dist},\mathcal{E}_{cls}^{s},\mathcal{E}_{cls}^{d},\mathcal{E}_{rec}$ according to formulas (4), (5), (6) and (9) respectively
   \STATE Optimize five networks $G$, $D_{di}$, $C_{di}$, $D_{ds}$, $C_{ds}$ according to formulas (7)
   \STATE Optimize network $F_{D}$ according to formulas (8)
   \STATE Optimize network $R$ according to formulas (9)
   \IF{$n$ \% $r$==1}
   \STATE Update two networks $C_{s}$ and $C_{t}$ according to formulas (10) and (11) respectively
   \ENDIF
   \STATE optimize network $G$  according to formulas (12)
   \ENDFOR
   \STATE \textbf{return} $G$,$D_{di}$,$C_{di}$
   \end{algorithmic}
   \end{algorithm}

In order to make the feature extraction network $G$ learn more compact features,
we first construct prototypes for each category in the source and target domains as mentioned above. Then
the prototypes are used to supervise the feature extraction network,
so the feature extraction network can extract original features which are close to the corresponding prototypes.
If the extracted feature of an image is closer to the corresponding prototype, then the features of each category in the source domain and the target domain will be more compact.
For the images in the source domain, the labels are available; while for the images in the target domain, the pseudo labels with high confidence calculated by $C_{di}$ are used. The optimization objective for $G$ are defined as
follows,
\begin{eqnarray}
   &&\min _{G} \theta\mathcal{E}_{g}= \theta(\mathcal{E}_{g}^{s}+\gamma \mathcal{E}_{g}^{t})
\end{eqnarray}
where $$\mathcal{E}_{g}^{s}=\frac{1}{n_{s}} \sum_{i=1}^{n_{s}}\mathcal{L}^{s}\left(C_{s}\left(f_{g}^{s}\right), y_{i}^{s}\right),$$ and
$$\mathcal{E}_{g}^{t}=\frac{1}{m}\sum_{i=1}^{m}\mathcal{L}^{s}\left(C_{t}\left(f_{g}^{t}\right), pseudo\right),$$ in which $\gamma$ is a hyperparameter for trading off and is set to 1 in all experiments which is consistent with the methods in \cite{xie2018learning},
and $m$ is the number of samples we selected based on the pseudo labels. The $\mathcal{E}_{g}$
is only applied on feature extraction network $G$, because we just want our
feature extraction network can learn more compact features. A hyperparameter $\theta$ is used for trading off of the optimization effect between Eqs. (7) and (12) which is set to 0.05 in most experiments based on cross-validation method.

Our method is summarized as the Algorithm DTR, and $C_{di}$ is chosen as the final classifier.
But what we can find is that we can also use $C_{t}$ to classify
the samples in the target domain.
Specifically, we can use the classifier $C_{t}$ to classify the original extracted feature $f_{g}^{t}$, or use the classifier $C_{di}$ to classify the domain invariant feature $f_{di}^{t}$.
In Section 4.3, we will compare these two classifiers and analyze the results.

\subsection{Analysis}
In this section, we theoretically show that our approach improves the boundary of the expected error on the target
samples is minimized by using the theory of domain adaptation in \cite{Bendavid}, and it also shows domain invariant features will be better extracted by more compact original features.
Formaly, let $\mathcal{H}$ be the hypothesis class.
Given two domains $\mathcal{S}$ and $\mathcal{T}$, the theory bounds the expected error of hypothesis $h$ on the
target samples $\mathcal{E}_{\mathcal{T}}(h)$ by three terms as follows:
\begin{equation}
  \forall h \in \mathcal{H}, \mathcal{E}_{\mathcal{T}}(h) \leq\mathcal{E}_{\mathcal{S}}(h)+\frac{1}{2}d_{\mathcal{H}\Delta\mathcal{H}}(\mathcal{S},\mathcal{T})+Con
\end{equation}
where $\mathcal{E}_{\mathcal{S}}(h)$ is the expected error of hypothesis $h$ on the source samples,
$d_{\mathcal{H}\Delta\mathcal{H}}(\mathcal{S},\mathcal{T})$  is the domain divergence
which measured between two distributions $\mathcal{S}$ and $\mathcal{T}$ with respect to a hypothesis set $\mathcal{H}$ and
$Con$ is the shared expected loss, which is usually considered as a constant.

In inequality (13), $\mathcal{E}_{\mathcal{S}}(h)$ can be minimized easily with source labeled data,
and $Con$ is usually considered as a constant, which is expected to be negligibly small.
Therefore, like many recent methods\cite{saito2018maximum,chen2019transferability}, our main consideration is how to minimize $\mathcal{E}_{\mathcal{T}}(h)$ by minimizing $d_{\mathcal{H}\Delta\mathcal{H}}(\mathcal{S},\mathcal{T})$.

Our core contribution lies in technically building prototypes and using them to learn more compact features.
From the perspective of features, we make the discrepancy between the original features and the prototype smaller according to the Eq.(12). According to our hypothesis, this discrepancy can be further rewritten as:
\begin{equation}
   \begin{aligned}
   \min \mathcal{E}_{g} &\Leftrightarrow \min \mathcal{L}^{s}\left(C\left(f_{g}\right), label\right) \\
   & \Leftrightarrow \min dis(f_{g},prototype) \\
   & \Leftrightarrow \min dis(f_{ds}\circ f_{di},w_{d}\circ w_{c}) ,
   \end{aligned}
\end{equation}
where $C$ represents classifier $C_{s}$ or $C_{t}$ and $dis(A,B)$ means the discrepancy between $A$ and $B$. $w_{d}$ and $w_{c}$ are domain prototype and class prototype respectively.
The reason why $\min \mathcal{L}^{s}\left(C\left(f_{g}\right), label\right) \Leftrightarrow \min dis(f_{g},prototype) $ is that when we optimize the network according to the Eq.(12), the weights of $C$ does not change, and $C$ is a linear classifier without bias, so $\min \mathcal{L}^{s}\left(C\left(f_{g}\right), label\right)$ is equivalent to requiring the extracted feature $f_{g}$ from the feature extraction network to be close to the weights of $C$, so as to reduce the discrepancy between $f_{g}$ and the prototype.

It is worth noting that in the process of domain classification, due to our domain label are available, so the supervised domain classification will make $f_{ds}$ and $w_{d}$ almost same.
So when we minimize the discrepancy between the prototype and the original feature, it is almost equivalent to minimize the distance between $f_{di}$ and $w_{c}$.
\begin{equation}
   \begin{aligned}
    \min dis(f_{ds}\circ f_{di},w_{d}\circ w_{c})\approx \min dis(f_{di}, w_{c}).
   \end{aligned}
\end{equation}

Because the class prototypes $w_{c}$ are shared in the source domain and the target domain, when we optimize and minimize the discrepancy between $f_{di}$ and $w_{c}$, it will naturally also minimize the discrepancy between $f_{di}^{s}$ and $f_{di}^{t}$,
so $d_{\mathcal{H}\Delta\mathcal{H}}(\mathcal{S},\mathcal{T})$ is minimized. Then it can further minimize the $\mathcal{E}_{\mathcal{T}}(h)$, which proves the effectiveness of our method.



\section{Experiments}
In this section, we first introduce our experimental setup.
Then, our experimental results are presented and compared with other state-of-the-art methods.
Finally, complete analysis are conducted for the proposed approach.

\subsection{Setup}

{\bf Digits}\cite{ganin2016domain-adversarial}
is a popular benchmark for visual domain adaptation. We use three digits datasets: MNIST, USPS, and SVHN.
And we evaluate our methods on three transfer tasks: MNIST to USPS(M$\rightarrow$U),
USPS to MNIST(U$\rightarrow$M), SVHN to MNIST(S$\rightarrow$M).

\textbf{Office-31}\cite{saenko2010adapting} is another popular benchmark for visual domain adaptation
that contains 4,652 images of 31 office environment categories from three domains: Amazon (A), DSLR (D) and
Webcam (W), which correspond to online website, digital
SLR camera and web camera images respectively.
And we evaluate our methods on all six transfer tasks.

\textbf{VisDA-2017}\cite{peng2017visda:} is a challenging benchmark for domain
adaptation. The source domain contains 152,397 synthetic images
that are renderings of 3D models.
The target domain has 55,388 real object images.

\textbf{Implementation Details}:
We implement our experiments on  Pytorch platform.
Following the standard evaluation
protocols for UDA, all labeled source and unlabeled target
samples are used as training data.
For fairer comparison with other methods
the setting we
used are same as \cite{long2018conditional} and \cite{chen2019transferability} for all tasks. Specifically,
for all tasks of Digits, the basic framework and optimizer we used are same as \cite{long2018conditional};
for all tasks  of office-31, we apply ResNet-50\cite{russakovsky2015imagenet} as the basic framework, and
the optimizer we used follows the paper \cite{chen2019transferability};
for the task  of VisDA-2017, we apply ResNet-101\cite{russakovsky2015imagenet} as the basic framework, and
the optimizer we used also follows the paper \cite{chen2019transferability}.
In all task, all disentanglers and reconstructor are two-layer fully connected networks and all classifiers are linear networks without bias.
The hyperparameters
$\alpha$ and $\gamma$ are set to 1 in all experiments refered to previous methods.
In most experiments, $\beta$ is set to 0.15 and $\theta$ is set to 0.05 based on cross-validation method.
\begin{table}[t]
   \centering
   \caption{Accuracy (\%) on Digits for domain adaptation.}
   \begin{tabular}{lccccccc}
   \hline

   Method & M$\rightarrow$ U & U$\rightarrow$ M & S$\rightarrow$ M & Avg \\ \hline
   DANN\cite{ganin2016domain-adversarial} & 90.4 & 94.7 & 84.2 & 89.8 \\ \hline
   ADDA\cite{tzeng2017adversarial} & 89.4 & 90.1 & 86.3 & 88.6 \\ \hline
   UNIT\cite{zhu2017unpaired} & 96.0 & 93.6 & 90.5 & 93.4 \\ \hline
   CDAN\cite{long2018conditional} & 93.9 & 96.9 & 88.5 & 93.1 \\ \hline
   CDAN+E\cite{long2018conditional} & 95.6 & 98.0 & 89.2 & 94.3 \\ \hline
   BSP+CDAN\cite{chen2019transferability} & 95.0 & 98.1 & 92.1 & 95.1 \\ \hline
   DTR(Proposed) & \bf{96.1}\bf{$\pm$0.4} & \bf{98.4}\bf{$\pm$0.5} & \bf{94.3}\bf{$\pm$0.8} & \bf{96.3} \\

   \hline

  \end{tabular}
  \end{table}
\begin{table*}[t]
 \centering
 \caption{Accuracy (\%) on Office-31 for domain adaptation.}
  \setlength{\tabcolsep}{4.0mm}
 \begin{tabular}{lccccccc}
 \hline

 Method & A$\rightarrow$W & D$\rightarrow$W & W$\rightarrow$D & A$\rightarrow$D & D$\rightarrow$A & W$\rightarrow$A & Avg \\ \hline
 ResNet-50\cite{he2016deep} & 68.4 & 96.7 & 99.3 & 68.9 & 62.5 & 60.7 & 76.1 \\ \hline
 DAN\cite{long2015learning} & 80.5 & 97.1 & 99.6 & 78.6 & 63.6 & 62.8 & 80.4 \\ \hline
 DANN\cite{ganin2016domain-adversarial} & 82.0 & 96.9 & 99.1 & 79.7 & 68.2 & 67.4 & 82.2 \\ \hline
 JAN\cite{long2017deep} & 85.4 & 97.4 & 99.8 & 84.7 & 68.6 & 70.0 & 84.3 \\ \hline
 GTA\cite{sankaranarayanan2018generate} & 89.5 & 97.9 & 99.8 & 87.7 & 72.8 & 71.4 & 86.5 \\ \hline
 CDAN\cite{long2018conditional} & 93.1 & 98.2 & \bf{100.0} & 89.8 & 70.1 & 68.0 & 86.6 \\ \hline
 CDAN+E\cite{long2018conditional} & 94.1 & \bf{98.6} & \bf{100.0} & 92.9 & 71.0 & 69.3 & 87.7 \\ \hline
 SAFN+ENT\cite{xu2019larger} & 90.1 & \bf{98.6} & 99.8 & 90.7 & 73.0 & 70.2 & 87.1 \\ \hline
 BSP+CDAN\cite{chen2019transferability} & 93.3 & 98.2 & \bf{100.0} & 93.0 & \bf{73.6} & 72.6 & 88.5 \\ \hline
 DTR(Proposed) & \bf{94.8$\pm$0.3} & 98.4$\pm$0.1 & 99.8$\pm$0.2 & \bf{93.8$\pm$0.2} & 73.1$\pm$0.3 & \bf{74.1$\pm$0.4} & \bf{89.0} \\ \hline

 \end{tabular}
 \end{table*}

\begin{table*}[t]
    \centering
    \caption{Accuracy (\%) on VisDA-2017 for domain adaptation.}
    \begin{tabular}{lccccccccccccc}
     \hline
     Method     & plane & bcybl & bus  & car  & horse & knife & mcyle & person & plant & sktbrd & train & truck & mean \\ \hline
     ResNet-101\cite{he2016deep} & 55.1  & 53.3  & 61.9 & 59.1 & 80.6  & 17.9  & 79.7  & 31.2   & 81.0  & 26.5   & 73.5  & 8.5   & 52.4 \\ \hline
     DAN\cite{long2015learning}        & 87.1  & 63.0  & 76.5 & 42.0 & 90.3  & 42.9  & 85.9  & 53.1   & 49.7  & 36.3   & \bf{85.8}  & 20.7  & 61.1 \\ \hline
     DANN\cite{ganin2016domain-adversarial}       & 81.9  & \bf{77.7}  & 82.8 & 44.3 & 81.2  & 29.5  & 65.1  & 28.6   & 51.9  & 54.6   & 82.8  & 7.8   & 57.4 \\ \hline
     MCD\cite{saito2018maximum}        & 87.0  & 60.9  & 83.7 & 64.0 & 88.9  & 79.6  & 84.7  & 76.9   & 88.6  & 40.3   & 83.0  & 25.8  & 71.9 \\ \hline
     CDAN\cite{long2018conditional}      & 85.2  & 66.9  & 83.0 & 50.8 & 84.2  & 74.9  & 88.1  & 74.5   & 83.4  & 76.0   & 81.9  & 38.0  & 73.7 \\ \hline
     BSP+CDAN\cite{chen2019transferability}   & 92.4  & 61.0  & 81.0 & 57.5 & 89.0  & 80.6  & 90.1  & 77.0   & 84.2  & \bf{77.9}   & 82.1  & 38.4  & 75.9 \\ \hline
     SAFN\cite{xu2019larger}   & \bf{93.6}  & 61.3  & \bf{84.1} & \bf{70.6} & \bf{94.1}  & 79.0  & \bf{91.8}  & \bf{79.6}   & \bf{89.9}  & 55.6   & 89.0  & 24.4  & 76.1 \\ \hline
    DTR(Proposed)       & 87.6    & 66.8   & 67.2 & 62.3 & 89.4  & \bf{90.0}  & 90.8  & 76.2   & 86.0  & 73.4   & 79.8  & \bf{51.5}  & \bf{76.8} \\

    \hline
    \end{tabular}
    \end{table*}

\subsection{Results}
The methods we compare with are DAN\cite{long2015learning},
DANN\cite{ganin2016domain-adversarial},
ADDA\cite{tzeng2017adversarial},
JAN\cite{long2017deep},
UNIT\cite{zhu2017unpaired},
GTA\cite{sankaranarayanan2018generate},
MCD\cite{saito2018maximum},
CDAN\cite{long2018conditional},
SAFN\cite{xu2019larger},
BSP+CDAN\cite{chen2019transferability}.
The classification accuracies on Digits, Office-31 and VisDA-2017
are shown in Tables 1-3 respectively.
As shown in Tables 1-3, our method is significantly better than the
state-of-the-arts in all datasets.

For Digits experiments, our method significantly outperforms the state-of-the-arts throughout all
experiments.
Especially in the experiment of S$\rightarrow$M, the performance is improved by 2.2\% compared with the best method in the past,
and for M$\rightarrow$U and U$\rightarrow$M, the performances are improved by 0.1\% and 0.3\% respectively compared with the best method in the past.

For Office-31 experiments, our method still yields start-of-the-art results overall.
Compared with the best results before,
the performances are improved by 0.7, 0.8, 1.5
in experiments A$\rightarrow$W, A$\rightarrow$D, and W$\rightarrow$A respectively.
For other experiments D$\rightarrow$W, W$\rightarrow$D, D$\rightarrow$A, we are not far behind the best current results.
For fairer comparison with those methods\cite{long2018conditional},
we used ten-crop images at the evaluation phase.

For VisDA-2017 experiment, the performance is improved by 0.7 in average compared with the best method in the past,
and from the accuracy rate of each category, we found that a substantial improvement has been
generated in the truck category.

The reason that our method can produce better classification results can be summarized as the following. When the gap between two domains is small,
the domain invariant feature can be more easily extracted from the original feature, which can help to find the better class prototypes such that the final extracted feature can be more discriminative and compact. So the classification model will work well.
When the gap between two domains is very large,
the domain specific feature can be more easily extracted from the original feature, which can help to find the better domain prototypes such that the better domain specific features can be obtained. Disentangling domain specific features  which have large differences between domains will obtain more better domain invariant features.
In summary, domain invariant and specific features in the disentanglement process
help each other to construct accurate prototypes to generate compact and discriminative features, which makes our classification model work better.


\begin{figure*}[t]
   \centering
       \subfigure[ResNet]{\includegraphics[width=0.22\linewidth]{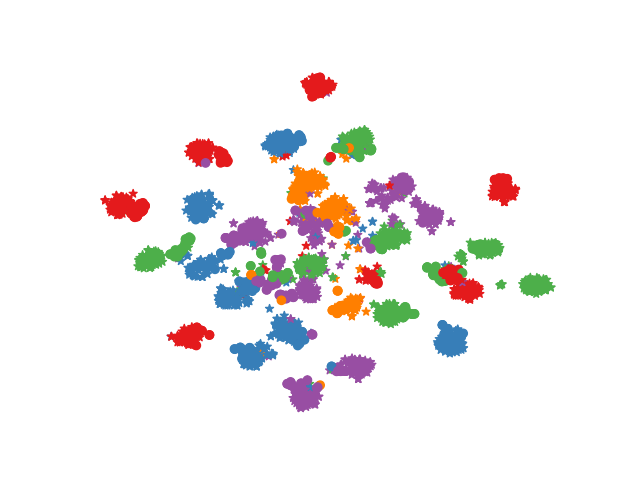}}
       \subfigure[DANN]{\includegraphics[width=0.22\linewidth]{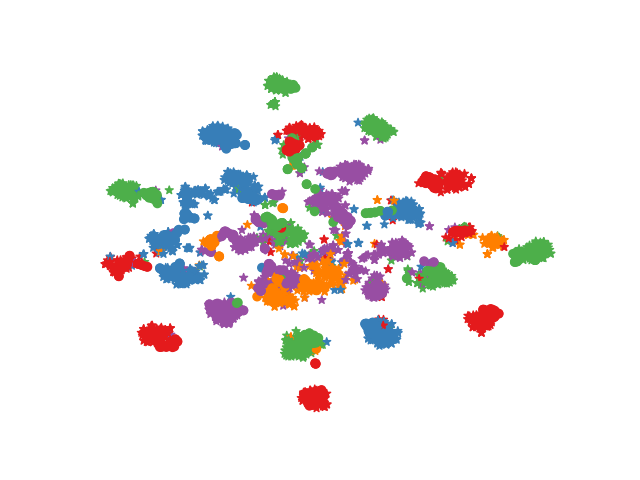}}
       \subfigure[DTR-$G$]{\includegraphics[width=0.22\linewidth]{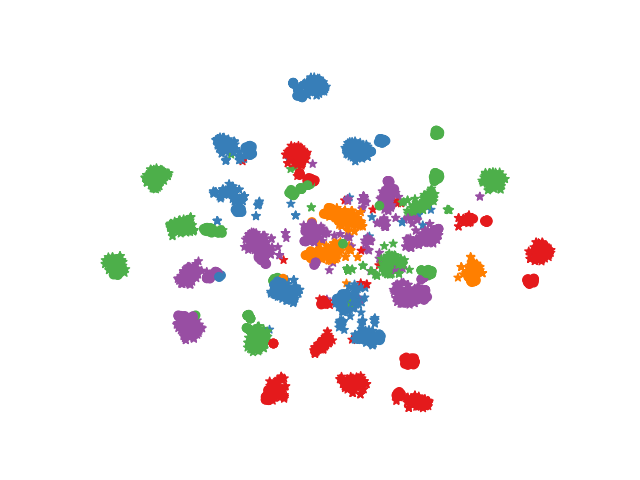}}
       \subfigure[DTR-$D_{di}$]{\includegraphics[width=0.22\linewidth]{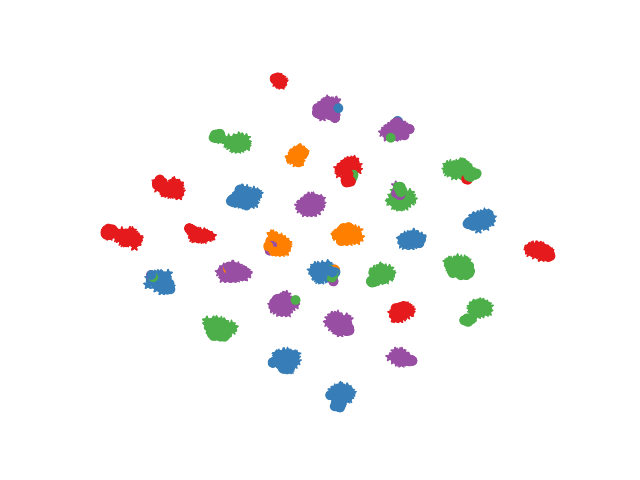}}

          \caption{Feature visualizations by T-SNE. The feature of task A$\rightarrow$W is learned by ResNet which only source data are used(a), DANN(b), feature extraction network $G$ trained by DTR(c), disentangler $D_{di}$ trained by DTR(d). }
 \end{figure*}

\subsection{Analyses}
In this section, several experiments are evaluated to further analyze our method.
First the classification results of the
$C_{di}$ and $C_{t}$ classifiers in the target domain are compared and analyzed.
Similarly, the classification results of $C_{ds}$ and $C_{s}$ in the source domain are also compared.
Second we investigate the sensitivity of the reconstruction interval $r$ as
it plays a significant role in reconstruction process.
Third, we visualize the features by t-SNE for an intuitive understanding.
And then we also measure $\mathcal{A}$-distance to measure the distribution discrepancy to further verification of our method.
Finally, we conduct a ablation study to investigate the impact of each of our parts in DTR.

\textbf{Comparisons between $C_{di}$ and $C_{t}$}: In this experiment,
based on Digits datasets the classification results
using $C_{di}$ and $C_{t}$ are compared in the target domain,
which are shown in Table 4. It is easy to found that almost all classification results of $C_{t}$ are worse than $C_{di}$.
This is because, even in a domain, there exist some samples,
whose domain specific features $f_{ds}$ are different from most of the samples in the domain.
That is, from the perspective of domain specific feature $f_{ds}$,
these samples are outlier.
As shown in Section 3.2,  $C_{t}$ classifies the combined features $f_{ds}^{t} \circ f_{di}$.
The performance of the classifier $C_{t}$ is degraded because of the large differences of $f_{ds}$ of some samples.
Therefore, the classification performance of $C_{t}$ is not as good as that of $C_{di}$.

To confirm our point, the classification results between $C_{di}$ and $C_{s}$ in the source domain are further compared.
The results are also shown in Table 4. Analyzing these results, we find that they are similar to the target domain, which confirms the rationality of our point of view.

 \begin{figure}[t]
   \begin{center}
      \includegraphics[width=0.8\linewidth]{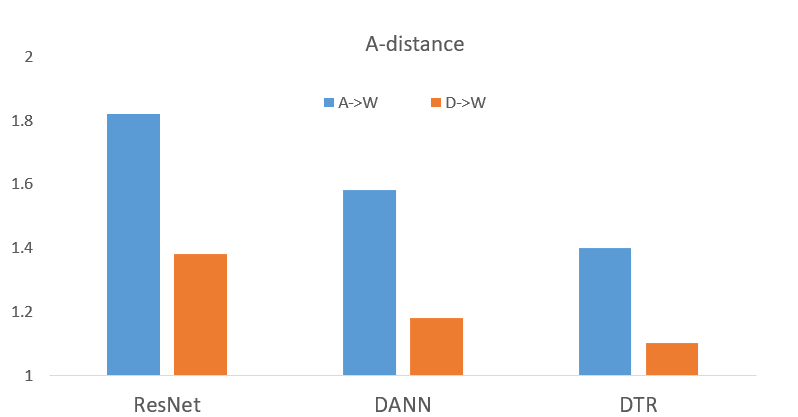}
   \end{center}
      \caption{Measure for distribution discrepancy. $\operatorname{dist}_{\mathcal{A}}$ on task A$\rightarrow$W, D$\rightarrow$W are showed with the features learned by ResNet, DANN, DTR.}
   \end{figure}

\begin{table}[t]
 \centering
 \caption{Comparisons between the classifiers $C_{di}$ and $C_{t}$ in the target domain (the first two rows), and comparisons between the classifiers $C_{di}$ and $C_{s}$ in the source domain (the last two rows).}
 \begin{tabular}{cccccccc}
 \hline
 Classifier & M$\rightarrow$U & U$\rightarrow$M & S$\rightarrow$M  \\ \hline
 $C_{t}$ & 93.4$\pm$0.5 & 97.4$\pm$0.5 & 94.3$\pm$0.7  \\ \hline
 $C_{di}$ & 96.1$\pm$0.4 & 98.4$\pm$0.5 & 94.3$\pm$0.8  \\ \hline
 $C_{s}$ & 97.7$\pm$0.3 & 97.1$\pm$0.2 & 95.5$\pm$0.4  \\ \hline
 $C_{di}$ & 99.9$\pm$0.1 & 99.2$\pm$0.2 & 98.2$\pm$0.2  \\ \hline
 \end{tabular}
\end{table}

\begin{table}[t]
 \centering
 \caption{Sensitivity analysis of reconstruction  interval $r$.}
 \begin{tabular}{cccccc}
 \hline

 The interval $r$ & 1 & 3 & 5 & 7 & 9 \\ \hline
 M$\rightarrow$U & 95.5 & 96.0 & 96.1 & 96.1 & 95.2 \\
                 & $\pm$0.8&$\pm$0.5&$\pm$0.4&$\pm$0.3&$\pm$0.3 \\ \hline
 U$\rightarrow$M & 97.7 & 98.2 & 98.4 & 98.4 & 98.0 \\
                 &$\pm$0.8&$\pm$0.6&$\pm$0.5&$\pm$0.3&$\pm$0.2\\ \hline
 S$\rightarrow$M & 93.8 & 94.3 & 94.3 & 94.0 & 93.6 \\
                 & $\pm$1.1 &$\pm$0.9&$\pm$0.8&$\pm$0.6&$\pm$0.6 \\ \hline
 \end{tabular}
 \end{table}

\textbf{Sensitivity of Reconstruction interval $r$}: The reconstruction interval $r$ plays an important role in reconstruction process.
If $r$ is small, the prototypes updated
during our disentanglement process are easily affected by outliers,
resulting in some noisy constructed original feature prototypes.
If $r$ is large, the original feature prototypes used to optimize feature
extraction network $G$ will be updated for a long time. The
constructed prototype used cannot accurately reflect the current optimal prototype because the update frequency is too
slow. So we need to take an appropriate intermediate value.
We conduct experiments on Digits transfer tasks which are
shown in Table 5. By fixing iteration number of batches, the
classification results increase first and then decrease as $r$ increases, which is in line with our analysis. Fortunately, the
value of $r$ does not affect the experimental results very much.

\textbf{Feature Visualization}: In this experiment, we visualize the feature of the task A$\rightarrow$W on Office-31 data by t-SNE. The features extracted by ResNet just trained in the source domain are shown
in Fig.3(a), and domain invariant features learned by DANN and DTR are shown in Fig.3(b) and
Fig.3(d) respectively. The features extracted by the network $G$ trained in both domains are shown in Fig.3(c).
By obsevering Fig.3(b) and Fig.3(d),
it can be clearly seen that the features extracted by DTR
are more concentrated. It proves that our method indeed extracts the compact features successfully.
By observing Fig.3(a) and Fig.3(c),
we can see that the discriminability of the extracted features is increasing,
which reflects the effectiveness of the reconstruction process.

\textbf{Distribution Discrepancy}: In this experiment, we use $\mathcal{A}$-distance
to measure the distribution discrepancy,
which defined as $\operatorname{dist}_{\mathcal{A}}=2(1-2 \epsilon)$ and $\epsilon$
is the test error of a classifier trained to classify the samples.
Fig. 4 shows $\operatorname{dist}_{\mathcal{A}}$ on tasks A$\rightarrow$W, W$\rightarrow$D
with features of ResNet, DANN, and DTR. And we can observe that the
$\operatorname{dist}_{\mathcal{A}}$ on DTR is smaller than $\operatorname{dist}_{\mathcal{A}}$
on both ResNet and DANN features, which implies that DTR can learn more domain invariant features than
both ResNet and DANN features. And we also can observe that the
$\operatorname{dist}_{\mathcal{A}}$ of task W$\rightarrow$D is smaller than that of task
A$\rightarrow$W. It implies that the features learned in task W$\rightarrow$D are more domain invariant and $C_{di}$ can achieve higher accuracy than
task A$\rightarrow$W. This is because the images between W and D are more similar than that between A and W, which can be clearly seen by observing the datasets.

\textbf{Ablation study}: In this experiment,
we use digital datasets to discuss the impact of each part in DTR.
The main purpose is to verify the validity of our hypothesis that it is useful to make features more compact through prototypes. Our experimental results are shown in Table 6.

"B"(Basel) denotes the baseline in which only domain invariant feature is disentangled from original feature, and CDAN-E\cite{long2018conditional} is used as adversarial training method, so "B" is the same as method CDAN-E.
"D" represents a complete disentanglement process which domain invariant and domain specific features are all disentangled from original feature.
"D+R" represents a method which has complete disentanglement process and partial reconstruction process which just calculate $\mathcal{E}_{rec}$.
"DTR" is the complete algorithm. Compared with "D+R", "DTR" learns more compact features by using prototypes.

By comparing the results of "B" and "D", we can easily find that the result of "B" is better than that of "D".
Compared with "B", "D" also needs to calculate domain classification loss. However, this part of loss does not help the network learn better domain invariant features in the process of "D". On the contrary, for learning better domain invariant features, this part of loss is a noise, so we can find that the result of "B" is better than that of "D".

From Table 6, we can easily find that the results of "D" and "D+R" are the same. Since we hope that $R$ can learn the nonlinear mapping we mentioned earlier. Thus the reconstruction loss $\mathcal{E}_{rec}$ only optimizes the network $R$, which is not backpropagated to other networks. So the results of "D" and "D+R" are the same.

By comparing "DTR" and "D+R", we find that the performance of "DTR" is much better than "D+R". The biggest difference between "DTR" and "D+R" is that "DTR" uses prototypes to learn more compact  features. Then we can conclude that using prototypes to learn compact features is extremely effective, which also proves our work hypothesis.

\begin{table}
  \centering
 \caption{Ablation study.}
 \setlength{\tabcolsep}{3mm}
  \begin{tabular}{cccc}
  \hline
  method & M$\rightarrow$U & U$\rightarrow$M & S$\rightarrow$M \\ \hline
  B      & 95.6              & 98.0              & 89.2              \\ \hline
  D      & 91.8              & 94.0              & 87.2              \\  \hline
  D+R    & 91.8              & 94.0              & 87.2              \\  \hline
  DTR    & 96.1              & 98.4              & 94.3             \\ \hline
  \end{tabular}
  \end{table}

\section{Conclusion}
We proposed a new technical route for unsupervised domain adaptation.
That is to use a disentanglement process to learn class prototypes and domain prototypes
and use a feature reconstruction process to construct the original feature prototypes which can supervise feature extraction network to extract more compact features,
so as to make the category classifier perform better.
As far as we know, it is the first time to construct prototypes by feature reconstructor for unsupervised domain adaptation.
Experiment results on several public data sets demonstrate the efficacy of
the proposed method.


%




\ifCLASSOPTIONcaptionsoff
  \newpage
\fi



\bibliographystyle{IEEEtran}
\bibliography{references}
\end{document}